\documentclass{article}
\usepackage{ijcai19}

\usepackage{times}
\usepackage{soul}
\usepackage{url}
\usepackage[hidelinks]{hyperref}
\usepackage[utf8]{inputenc}
\usepackage[small]{caption}
\usepackage{graphicx}
\usepackage{amsmath}
\usepackage{booktabs}
\usepackage{cite}
\title{A Survey on Causal Representation Learning and Future Work for Medical Image Analysis}

\author{
Changjie Lu$^1$
\affiliations
$^1$Wenzhou-Kean University\\
\emails
lucha@kean.edu
}

\begin{document}

\maketitle

\begin{abstract}
Statistical machine learning algorithms have achieved state-of-the-art results on benchmark datasets, outperforming humans in many tasks. However, the out-of-distribution data and confounder, which have an unpredictable causal relationship, significantly degrade the performance of the existing models. Causal Representation Learning (CRL) has recently been a promising direction to address the causal relationship problem in vision understanding. This survey presents recent advances in CRL in vision. Firstly, we introduce the basic concept of causal inference. Secondly, we analyze the CRL theoretical work, especially in invariant risk minimization, and the practical work in feature understanding and transfer learning. Finally, we propose a future research direction in medical image analysis and CRL general theory.
\end{abstract}

\section{Introduction}



Correlation does not imply causation \cite{geer2011correlation}. One famous example is the Simpson's paradox\cite{wagner1982simpson} (see Fig. \ref{sp}). Even if the series of events do have causality, it is hard to distinguish that relationship. One effective way of learning causality is to conduct a randomized controlled trial (RCT), randomly assigning participants into a treatment group or a control group so that people can observe the effect via the outcome variable. However, RCT is inflexible because it targets the sample average, which makes the mechanism unclear. Another widely used information type is observational data, which records every event that could be observed. 
Nowadays, machine learning algorithms attempt to learn patterns by fitting the observational data, losing sight of the causality. This results in poor performance when generalizing the model to an unseen distribution or learning the wrong causality.

To escape the dilemma mentioned above, Pearl firstly introduced a causality system with the three-layer causal hierarchy, called Pearl Causal Hierarchy (PCH), which contains Association, Intervention, and Counterfactuals\cite{pearl2009causal}\cite{pearl2}\cite{pearl2009causality}. To support this theory, Pearl developed structural causal model (SCM), which combines structural equation models (SEM), potential outcome framework, and the directed acyclic graphs (DAG)\cite{description1}\cite{description2}\cite{description3}\cite{description4}\cite{description5} for probabilistic reasoning and causal analysis, typically using the do-calculus\cite{do1}.
With these tools, causal analysis infers probabilities under not only statistical conditions but also the dynamics of probabilities under changing conditions \cite{pearl2009causality}. Currently, causal inference is a popular research direction with comprehensive literature \cite{yao2021survey}\cite{imbens2015causal}\cite{gangl2010causal}\cite{bareinboim2022pearl}\cite{zhang2021causal}, and is widely applied in decision evaluation (e.g. healthcare), counterfactual estimation (e.g. representation learning methods), and dealing with selection bias (e.g. advertising, recommendation) \cite{yao2021survey}.

\begin{figure}
    \centering
    \includegraphics[width=5.5cm]{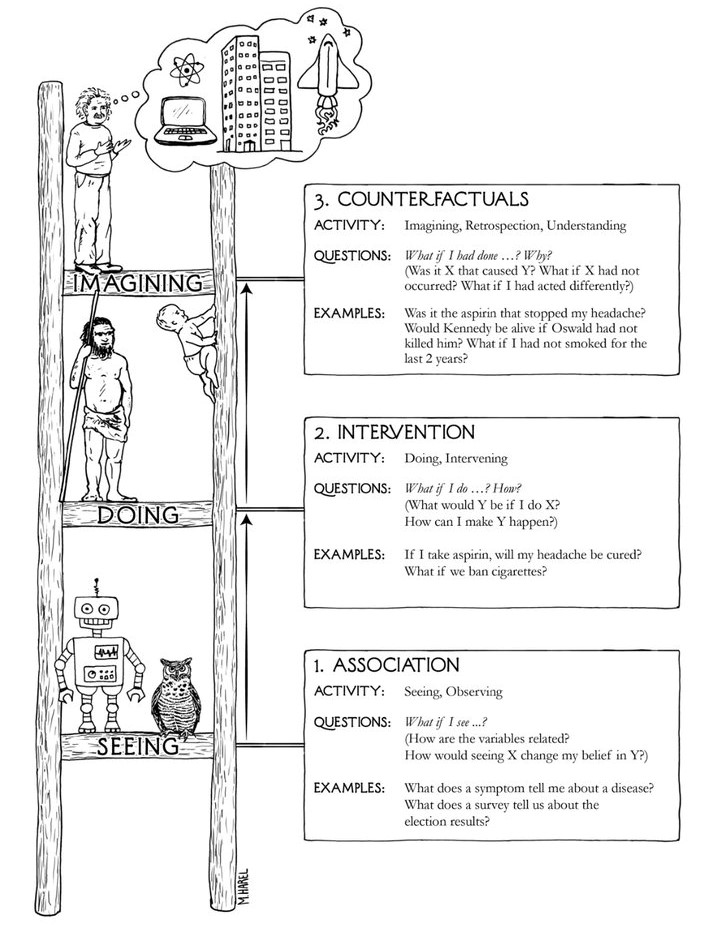}
    \caption{Three levels of Pearl Causal Hierarchy (PCH). (Drawing by \cite{picture}) The first level is association $P(y|x)$,(e.g. supervised or unsupervised learning). The second level is intervention $P(y|do(x),c)$, e.g.(feature learning, few-show learning). The third level is counterfactual $P(y_{x}|x',y')$, (e.g. zero-shot learning, long-tailed classification.)}
    \label{PCH}
\end{figure}

Although most of the causal inference could be applied in low-dimensional data (e.g., tabular data, describable events), research in high-dimensional data is still a struggle. In computer vision, for example, which often suffers from confounder that the model tends to classify a cow in dessert as a camel. The invention of advanced network architecture (i.e. resnet\cite{he2016deep},transformer\cite{vaswani2017attention}, etc.) may even enhance this misunderstanding. Recently, 
much research introduced the task-driven solution, attempting to discover the fundamental mechanism. (e.g. in image deraining, \cite{ren2019progressive}\cite{zheng2022sapnet}\cite{jiang2020multi} introduced the progressive algorithm, in low-light image enhancement, \cite{cuiyou} imitate the principle of camera imaging, in point cloud analysis, \cite{qi2017pointnet}\cite{qi2017pointnet++} introduced point abstraction.) However, these models still struggle in o.o.d prediction. 

Causal representation learning (CRL) is a useful tool to unscramble mechanisms. CRL assumes the data are latent causal variables that are causal related and satisfy conditional SCM, using non-linear mapping. With this assumption, CRL could discover the causal relationship via estimating the distribution after intervention if the causal latent variable and SCM are learnable. Moreover, CRL could even imagine the unseen data according to the counterfactual results, making models robust in the o.o.d prediction. However, distinguishing the confounder and discovering the SCM is very challenging. Therefore, some assumptions like sparsity and independent causal mechanism are introduced as an inductive bias in CRL\cite{scholkopf2021toward}. Recently, theoretical works with CRL have been developed (e.g. Low-rank\cite{low1}\cite{low2}, Generalized Independent Noise condition\cite{gin1}\cite{gin2}, invariant risk minimization\cite{d1-irm1,d1-irm2,d1-irm3,d1-irm4,d1-irm5,d1-irm6,d1-irm7,d1-irm8}) and has shown promising performance in feature understanding (e.g. scene graph generation, pretraining, long-tailed data)\cite{wang2020visual}\cite{tang2020unbiased}\cite{f5,f6,f7,f8,f9,f10,f11,f12,f13,f14,f15} and transfer learning problem (e.g. adversarial methods, generalization, adaptation)\cite{d2-domain1,d2-domain2,d2-domain3,subbaswamy2020spec,d2-domain4,d2-domain5}.

In this paper, we present a survey on CRL about its recent advances, with a special focus on the basic concept of causal inference (section 2), theoretical work, practical work(section 3), and future research directions in medical image analysis\cite{med3}(section 4).
\begin{figure*}[h]
    \centering
    \includegraphics[width=16cm]{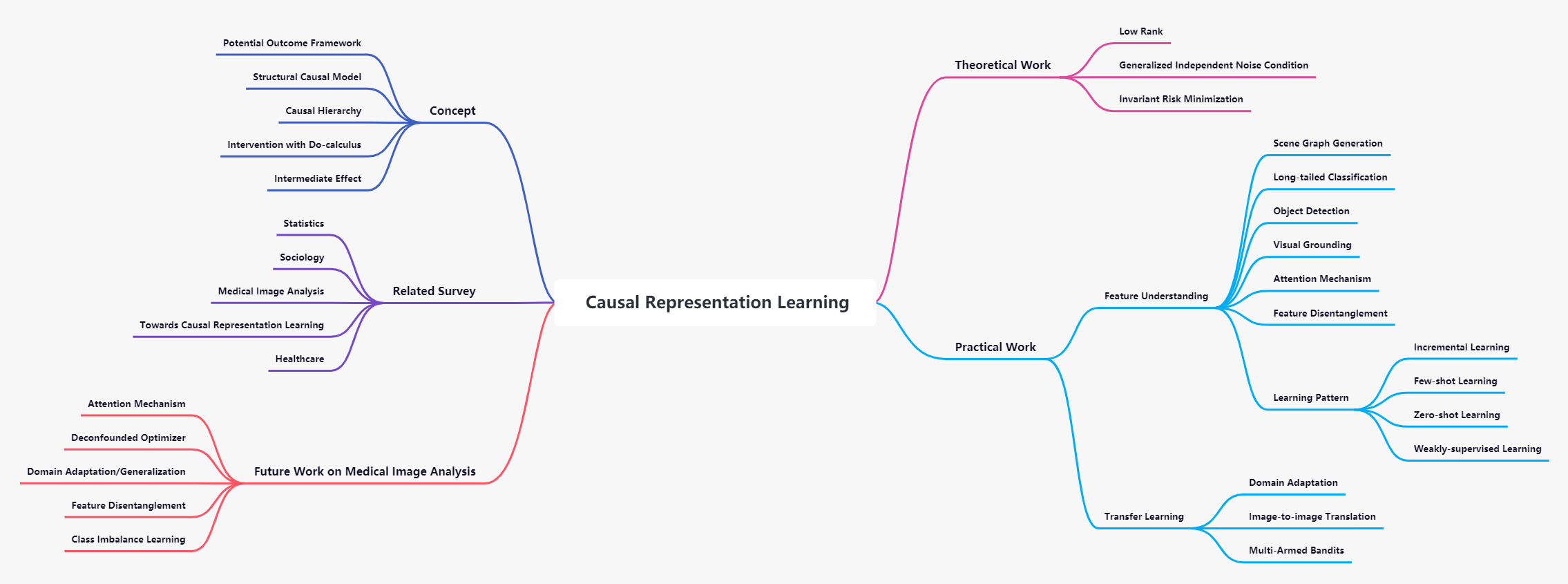}
    \caption{The mindmap of causal representation learning in vision}
    \label{mindmap}
\end{figure*}

\section{Concept of Causal Inference}
\begin{table}[h]
 \setlength{\tabcolsep}{1.5mm}{
\begin{tabular}{ccccl}
\hline
                               & \multicolumn{4}{c}{Condition}                                                                                                                                                                               \\ \hline
\multicolumn{1}{l|}{Treatment} & Mild                                                      & Severe                                                   & Total                                                              & Causal          \\ \hline
\multicolumn{1}{c|}{A}         & \begin{tabular}[c]{@{}c@{}}15\%\\ (210/1400)\end{tabular} & \begin{tabular}[c]{@{}c@{}}30\%\\ (30/100)\end{tabular}  & \textbf{\begin{tabular}[c]{@{}c@{}}16\%\\ (240/1500)\end{tabular}} & 19.4\%          \\ \hline
\multicolumn{1}{c|}{B}         & \begin{tabular}[c]{@{}c@{}}10\%\\ (5/50)\end{tabular}     & \begin{tabular}[c]{@{}c@{}}20\%\\ (100/500)\end{tabular} & \begin{tabular}[c]{@{}c@{}}19\%\\ (105/550)\end{tabular}           & \textbf{12.9\%} \\ \hline
\end{tabular}}
\caption{The example of Simpson's paradox. Although the total deaths of treatment A is less than deaths of treatment B, the treatment B is a worthy choice in terms of causal analysis. In naive method, for example, the treatment effect of A (16\%), is given by $\frac{1400}{1500}(0.15)+\frac{100}{1500}(0.30)$. In causal analysis, the treatment effect of A (19.4\%) is given by $\frac{1450}{2050}(0.15)+\frac{600}{2050}(0.30)$.}
\label{sp}
\end{table}
In this section, several concepts of causal reasoning are introduced, including potential outcome framework, structural causal model(SCM), and causal graphs via do-calculus.

\subsection{Potential outcome framework}
\begin{figure}[h]
    \centering
    \includegraphics[width=8cm]{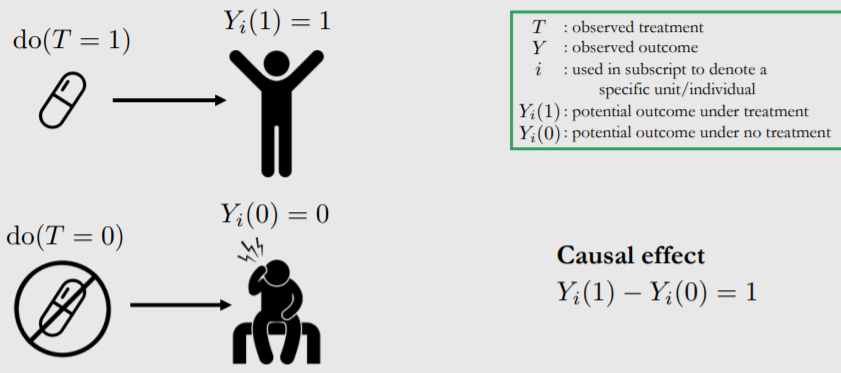}
    \caption{Example of the potential outcomes (Drawn by \cite{neal2020introduction}). The treatment effect are measured on whether to take the drug or not. }
    \label{pof}
\end{figure}
An example in Fig.\ref{pof} shows how the potential outcomes framework works. However, we are in a dilemma that we cannot observe the causal effect on the same person. If one person takes the drug, we will lose the information of the person not taking the drug. This dilemma is known as the "fundamental problem of causal inference"\cite{zhang2021causal}.
\subsection{Structural Causal Model}
Firstly, we define the symbols.
\begin{itemize}
    \item $X$, random variable
    \item $P\left(Y_x\right):=P(Y \mid d o(X=x))$
    \item path $(X,Y)$: any path from X to Y.
    \item collider $Z$, $X \rightarrow Z \leftarrow Y, X \perp Y, X \not\perp Y\mid Z$.
\end{itemize}
The structural causal model\cite{pearl2009causal}\cite{pearl2009causality} is a 4-tuple $\langle\mathbf{U}, \mathbf{V}, \mathcal{F}, P(\mathbf{U})\rangle$, where:
\begin{itemize}
    \item $\mathbf{U}$ is a set of background variables (exogenous variables), determined by factors outside the model.
    \item $\mathbf{V}$ is a set $\left\{V_1, V_2, \ldots, V_n\right\}$ of variables, called endogenous, determined by other variables within the model
    \item $\mathcal{F}$ is a set of functions $\left\{f_1, f_2, \ldots, f_n\right\}$, each $f_{i}$ is a mapping from $U_i \cup P A_i$(PA: parents) to $V_{i}$, where $U_i \subseteq U$ and $PA_{i}$ is a set of causes of $V_{i}$. The entire set $\mathcal{F}$ forms a mapping from $\mathbf{U}$ to $\mathbf{V}$. That is, for $i$ = 1,...,n, $v_i \leftarrow f_i\left(p a_i, v_i\right)$.
    \item $P(\mathbf{U})$ is a probability function defined over the domain of $\mathbf{U}$.
\end{itemize}
For example, suppose that there exists a causal relationship between treatment solution $X$ and lung function $Y$ of an asthma patient. 
Simultaneously, suppose that $Y$ also relies on the level of air pollution $Z$. Under this circumstance, $X$ and $Y$ are endogenous variables, $Z$ is exogenous variables. Therefore, the SCM can be instantiated as,
\begin{equation}
\begin{aligned}
        & U = \{Z,U_{x},U_{y}\},  V = \{X,Y\},  F = \{f_{X},f_{Y}\} \\
    & f_{X} : X \leftarrow f_{X}(U_{x}) \\
    & f_{Y} : Y \leftarrow f_{Y}(X,Z,U_{y})
\end{aligned}
\end{equation}
\subsection{Causal Hierarchy}
\subsubsection{Level1 seeing}
For any SCM, the formula,
\begin{equation}
    P^{\mathcal{M}}(\mathbf{Y}=\mathbf{y})=\sum_{\{\mathbf{u} \mid \mathbf{Y}(\mathbf{u})=\mathbf{y}\}} P(\mathbf{u})
\end{equation}
could estimate any joint distribution of $\mathbf{Y} \subset \mathbf{V}$ given by $\mathbf{Y}(U = u)$. Take the image classification task as an example. $\mathbf{V}=\mathbf{X} \bigcup \mathbf{Y}$, $X$ represents the images, $Y$ represents the labels. The aim is to model $P(\mathbf{Y}|\mathbf{X})$. At this level, we could only build the model by fitting the distribution of observational data.
\subsubsection{Level2 doing}
In the level of doing, a hypothesis is proposed and then verified. In this condition, a new SCM is built: $\mathcal{M}_{\mathbf{x}}=\left\langle\mathbf{U}, \mathbf{V}, \mathcal{F}_{\mathbf{x}}, P(\mathbf{U})\right\rangle$, where $\mathcal{F}_{\mathbf{X}}=\left\{f_i: V_i \notin \mathbf{X}\right\} \cup(\mathbf{X} \leftarrow \mathbf{x})$. Therefore, $P^{\mathcal{M}}$ can be estimated by,
\begin{equation}
    P^{\mathcal{M}}\left(\mathbf{Y}_{\mathbf{x}}=\mathbf{y}_{\mathbf{x}}\right)=\sum_{\left\{\mathbf{u} \mid \mathbf{Y}_{\mathbf{x}}(\mathbf{u})=\mathbf{y}_{\mathbf{x}}\right\}} P(\mathbf{u})
\end{equation}
where $\mathbf{Y}_{\mathbf{x}}(\mathbf{u})$ represents $\mathcal{F}_{\mathrm{x}}(U=u)$.
\subsubsection{Level3 imaging}
In the level of imaging, the target is to know the effect whether another decision had been made, which can be formulated by,
\begin{equation}
    P\left(Y_{x^{\prime}}^{\prime} \mid X=x, Y=y\right)
\end{equation}
Namely, imaging $d o\left(X=x^{\prime}\right)$ given $X = x, Y = y$. Based on this, the joint distribution $P^{\mathcal{M}}$ can be estimated by,
\begin{equation}
    P^{\mathcal{M}}\left(\mathbf{y}_{\mathbf{x}}, \cdots, \mathbf{z}_{\mathbf{w}}\right)=\sum_{\left\{\mathbf{u} \mid \mathbf{Y}_{\mathbf{x}}(\mathbf{u})=\mathbf{y}_{\mathbf{x}}, \cdots, \mathbf{Z}_{\mathbf{w}}(\mathbf{u})=\mathbf{z}\right\}} P(\mathbf{u})
\end{equation}
for any $\mathbf{Y}, \mathbf{Z}, \cdots, \mathbf{X}, \mathbf{W} \subset \mathbf{V}$.

\subsection{Intervention with do-calculus}
\subsubsection{D-separation}
\begin{figure}
    \centering
    \includegraphics[width=5cm]{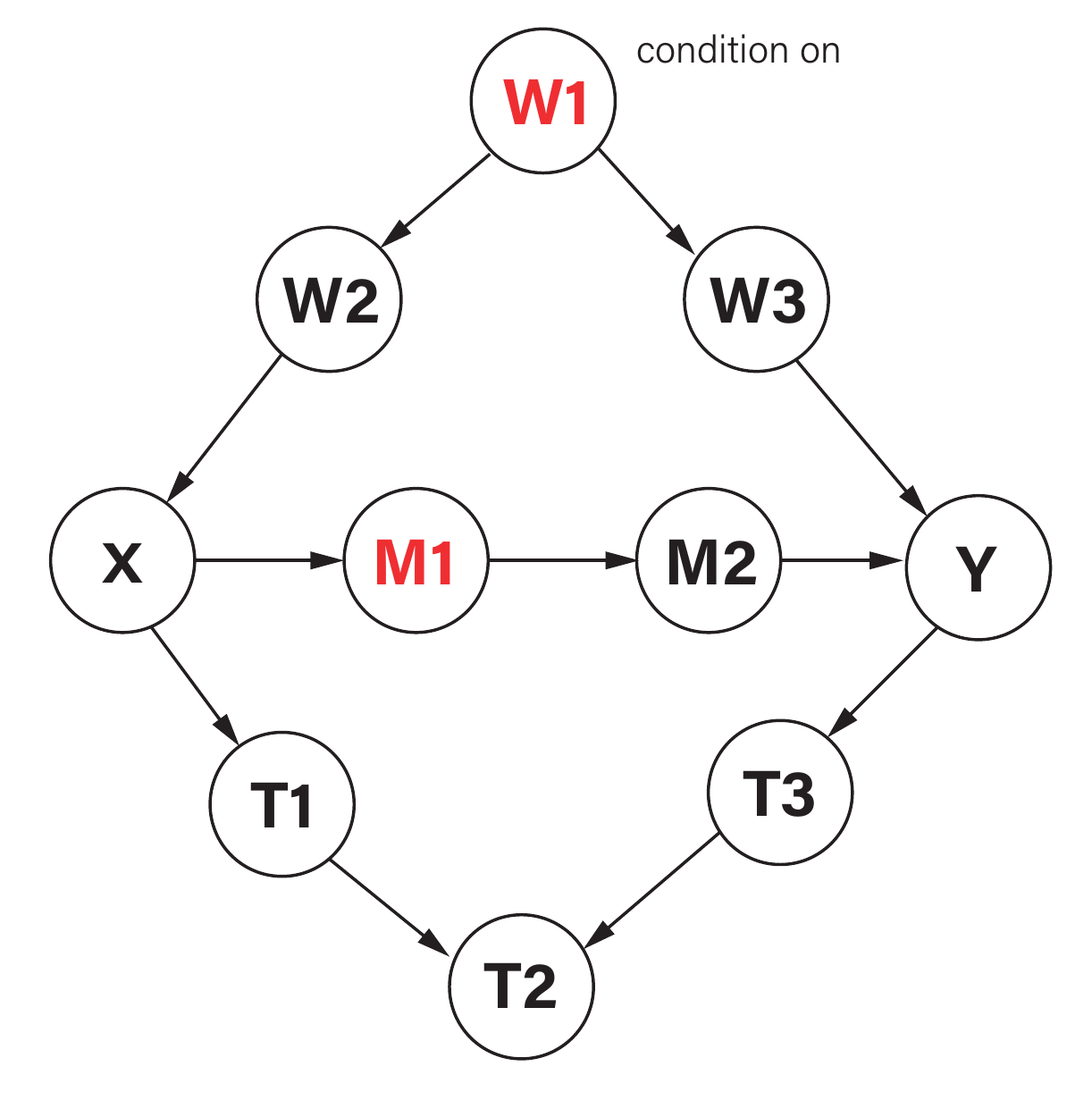}
    \caption{The demonstration for the d-separation. W1 and M1 are conditioned, which block all possible way from $X \rightarrow Y$. Therefore, X,Y are d-separated.}
    \label{s-p}
\end{figure}
Two sets of nodes $X$ and $Y$ are d-separated by a set of nodes $Z$ if all of the paths between any node in $X$ and any node in $Y$ are blocked by $Z$. In Fig. \ref{s-p}, we provide an example to illustrate d-separation. If $X$ is the cause, $Y$ is the effect. The other node is confounding. If W1/W2/W3 and M1/M2 are conditioned on, $X$ and $Y$ are d-separated. If T2 is conditioned on, $X$ and $Y$ are not d-separated because the relationship between $X$ and $Y$ could be found by intervening T2. If both T1 and T2 are conditioned on, $X$ and $Y$ are d-separated because T1 blocks the information path at the bottom of the figure. This concept explains why $X \rightarrow M1 \rightarrow M2 \rightarrow Y$ is causal association and $X \rightarrow W1 \rightarrow W2 \rightarrow W3 \rightarrow Y$, $X \rightarrow T1 \rightarrow T2 \rightarrow T3 \rightarrow Y$ are non-causal associations.
\subsubsection{Intervention}
\begin{figure}[h]
    \centering
    \includegraphics[width=8cm]{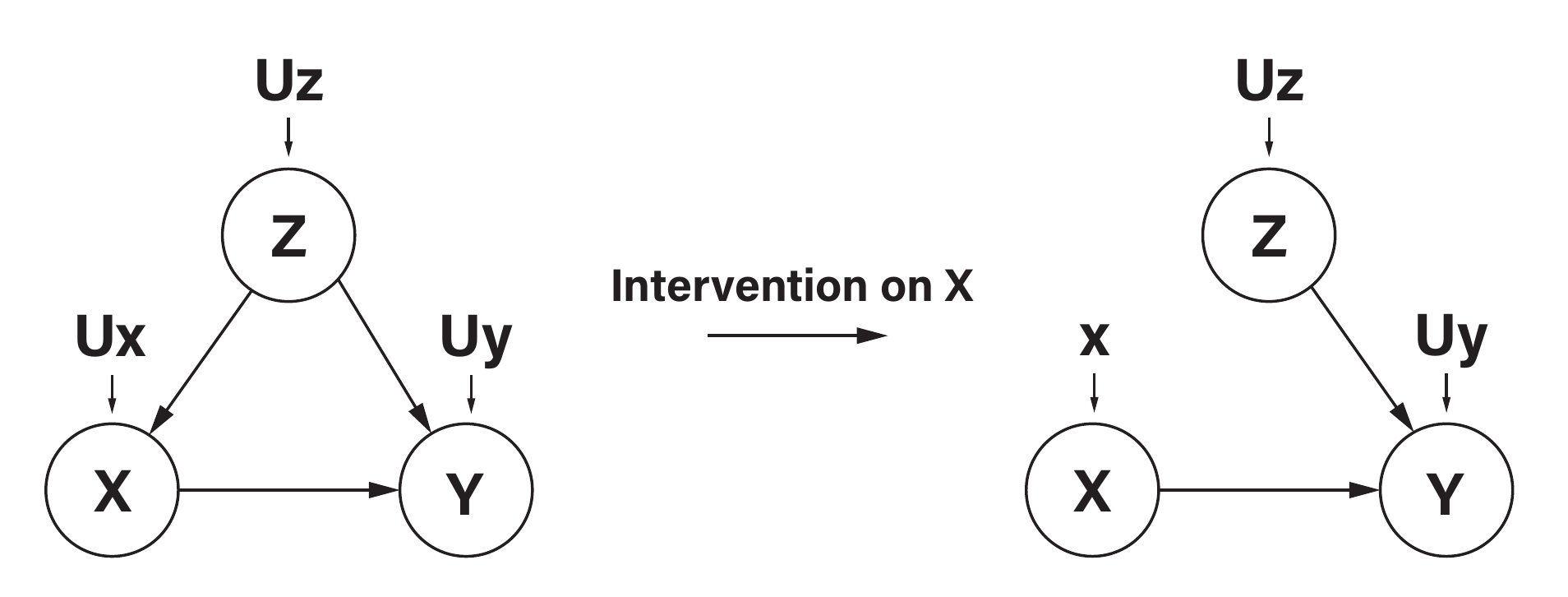}
    \caption{After the intervention on $X=x$, the edge from $Z \rightarrow X$ should be deleted.}
    \label{inven}
\end{figure}
We use $do(X=x)$ to represent intervention, $P(Y=y|do(X=x))$ represents the probability of $Y=y$ when making $X=x$. In the graphical model, one edge is deleted to represent the intervention on a particular node. From Fig.\ref{inven}, two invariant equations can be formulated by,
\begin{equation}
\begin{aligned}
    &P_{m}(Y=y|Z=z,X=x) = P(Y = y|Z=z,X=x) \\
    &P_{m}(Z=z) = P(Z = z)
\end{aligned}
\end{equation}
$Z$ and $X$ are d-separated after the modification, indicating that,
\begin{equation}
    P_{m}(Z=z|X=x)=P_{m}(Z=z)=P(Z=z)
\end{equation}
Therefore,
\begin{equation}
\begin{aligned}
      & P(Y=y|do(X=x)) = P_{m}(Y=y|X=x) \\
      & = \sum_{z}{P_{m}(Y=y|X=x,Z=z)P_{m}(Z=z|X=x)} \\
      & = \sum_{z}{P_{m}(Y=y|X=x,Z=z)P_{m}(Z=z)}
\end{aligned}
\end{equation}
Finally, the causal effect formula before intervention can be obtained using the relationship of invariance,
\begin{equation}
    P(Y=y|do(X=x)) = \sum_{z}P(Y=y|X=x,Z=z)P(Z=z)
\end{equation}
This formula is called the adjustment formula, calculating the relationship between $X$ and $Y$ For every value of $Z$.

Consider the example in Table.\ref{sp} and the causal graph in Fig.\ref{inven}. $X = A/B$ donates to patients who take the drug A/B. $Z = Mild/Severe$ donates the level of illness. $Y$ donates the death rate. \\
The effect of taking the drug A:
\begin{equation}
    E[Y|do(X=A)] = \frac{1450}{2050}(0.15) + \frac{600}{2050}(0.30) \approx 0.194
\end{equation}
The effect of taking the drug B:
\begin{equation}
    E[Y|do(X=B)] = \frac{1450}{2050}(0.10) + \frac{600}{2050}(0.20) \approx 0.129
\end{equation}
The result indicates that treatment B has a better effect which is contradictory to the statistical results.
\subsubsection{Backdoor criterion}
\begin{figure}
    \centering
    \includegraphics[width=7cm]{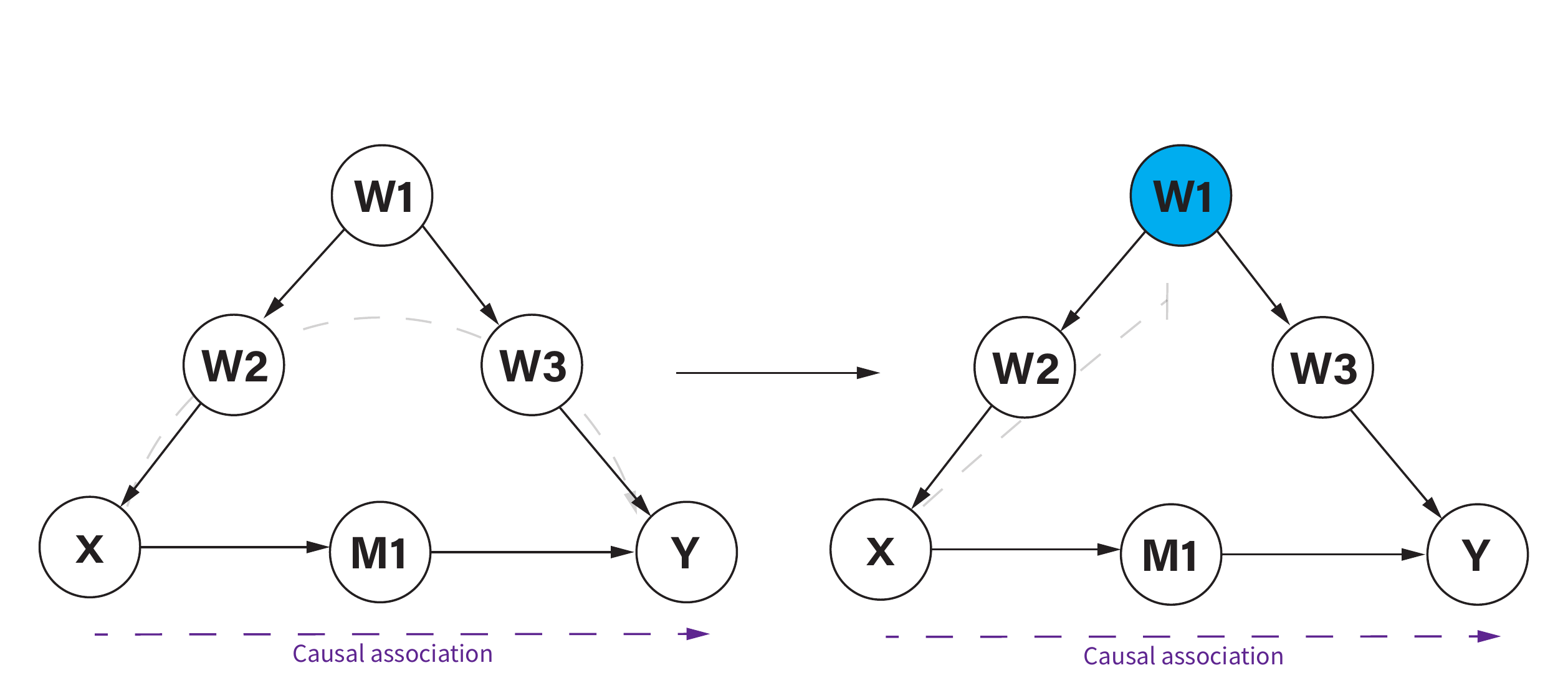}
    \caption{The example of the backdoor adjustment. The second line in Equ.\ref{ba}: W is a sufficient adjustment set, blocking all backdoor paths, only reserving the causation $X \rightarrow Y$. Third line in Equ.\ref{ba}: $do(X)$ blocks all T's parents.}
    \label{backadjust}
\end{figure}
\begin{figure}
    \centering
    \includegraphics[width=6cm]{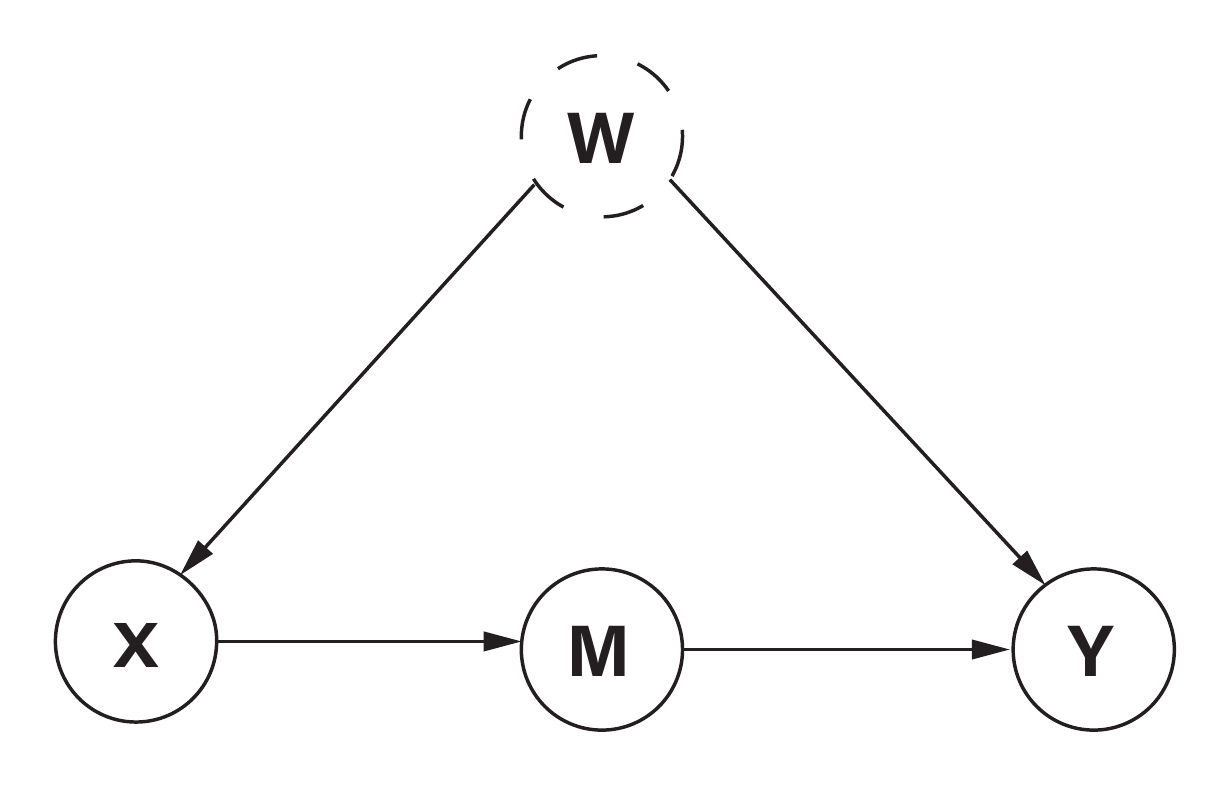}
    \caption{The example of the frontdoor adjustment. Typically, we do twice backdoor adjustments. Firstly, from $X \rightarrow M$, there is no backdoor path. Secondly, from $M \rightarrow Y$, T block the backdoor path $M \leftarrow X \leftarrow W \rightarrow Y$. Therefore, the final equation is defined in step 3.}
    \label{frontadjust}
\end{figure}
$W$ is said to satisfy the backdoor criterion about $(X,Y)$, if $W$:
\begin{itemize}
    \item blocks all paths between $X$ and $Y$.
    \item keeps the same directed paths from $X \rightarrow Y$.
    \item does not produce a new path.
\end{itemize}
\textbf{Backdoor adjustment}: if $W$ satisfies the backdoor criterion, then ATE (Average Treatment Effect) is identified.
\begin{equation}
    \begin{aligned}
P(y \mid d o(X)) &=\sum_w P(y \mid d o(X), w) P(w \mid d o(X)) \\
&=\sum_w P(y \mid X, w) P(w \mid d o(X)) \\
&=\sum_w P(y \mid X, w) P(w)
\end{aligned}
\label{ba}
\end{equation}

\subsubsection{Frontdoor criterion}
$W$ is said to satisfy the backdoor criterion about $(X,Y)$, if:
\begin{itemize}
    \item $W$ blocks all possible directed path from $X \rightarrow Y$.
    \item There is no backdoor path from $X \rightarrow W$
    \item All possible paths from $W \rightarrow Y$ are blocked by $X$.
\end{itemize}
\textbf{Frontdoor adjustment}:
\begin{itemize}
    \item X on M: $P(m \mid d o(x))=P(m \mid x)$
    \item M on Y: $P(y \mid d o(m))=\sum_x P(y \mid m, x) P(x)$
    \item X on Y: $P(y \mid d o(x))=\sum_m P(m \mid d o(x)) P(y \mid d o(m))=\sum_m P(m \mid x) \sum_{x^{\prime}} P\left(y \mid m, x^{\prime}\right) P\left(x^{\prime}\right)$
\end{itemize}
\subsubsection{Intermediate Effect}
\begin{figure}[h]
    \centering
    \includegraphics[width=6cm]{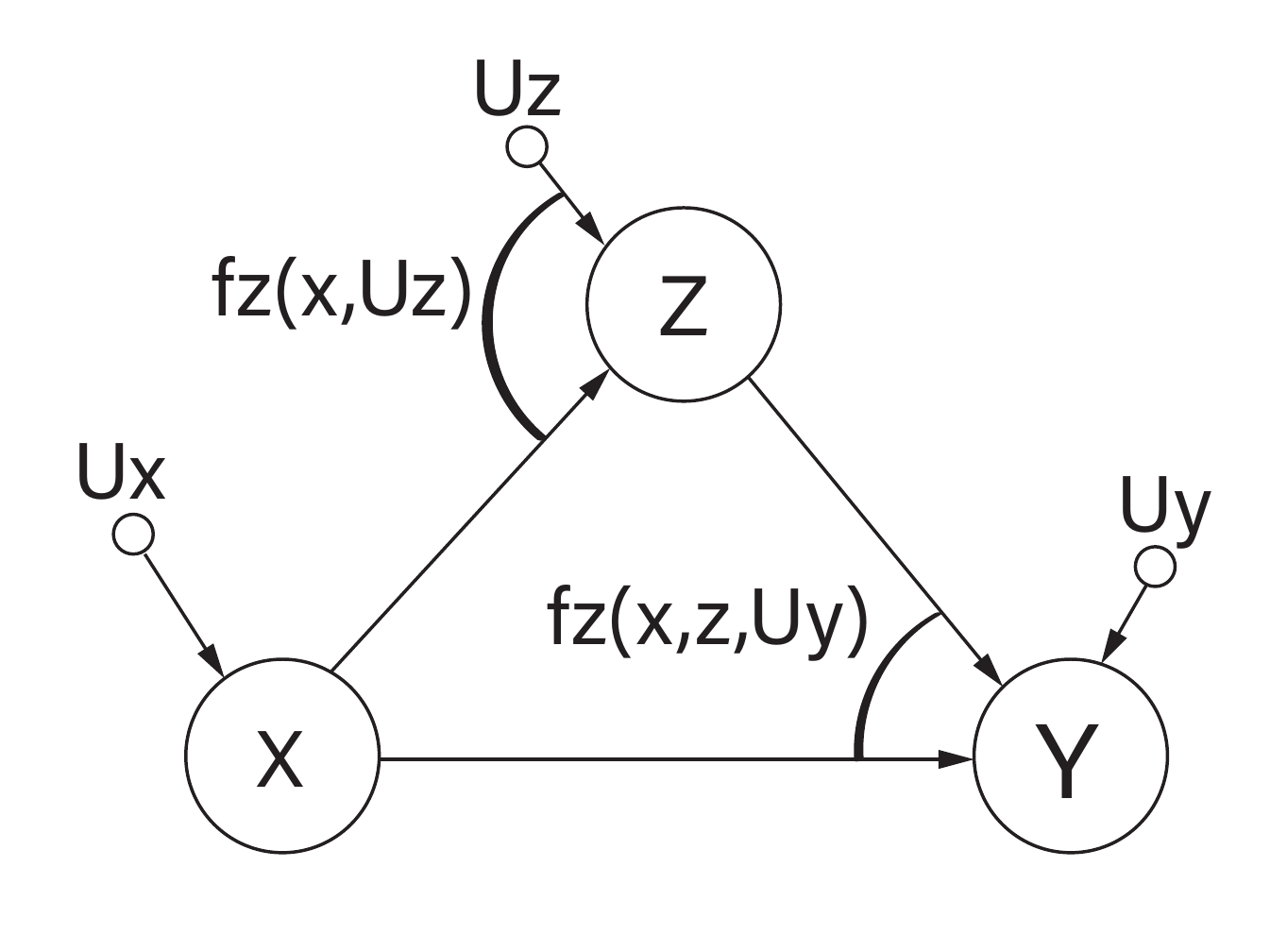}
    \caption{The mediation model without confounding.}
    \label{intermed}
\end{figure}
In a causal model, a classical intermediate problem can be defined as:
\begin{equation}
    x=f_X\left(U_X\right), z=f_Z\left(x, U_Z\right), y=f_Y\left(x, z, U_Y\right)
\end{equation}
where $X$ donates treatment, $Z$ donates mediator, $Y$ donates outcome. $f_X,f_Z,f_Y$ are any function. $U_X,U_Z,U_Y$ are background variables (see. Fig.\ref{intermed}).
Based on this, the effect is analyzed by the intervention as follows,\\
Average Treatment Effect:
\begin{equation}
    \begin{aligned}
          ATE& = E[Y_{1}] - E[Y_{0}]\\
          &= E[Y \mid do(X=1)]-E[Y \mid do(X=0)]] 
    \end{aligned}
\end{equation}
Controlled Direct Effect:
\begin{equation}
    \begin{aligned}
       &C D E(z)=E\left[Y_{1, Z}-Y_{0, Z}\right]\\
       &=E[Y \mid d o(X=1, Z=z)]-E[Y \mid d o(X=0, Z=z)] \\
    \end{aligned}
\end{equation}
Natural Direct Effect:
\begin{equation}
 N D E = E\left[Y_{1, Z_0}-Y_{0, Z_0}\right]
\end{equation}
Natural Indirect Effect:
\begin{equation}
     NIE = E\left[Y_{0, Z_1}-Y_{0, Z_0}\right]
\end{equation}
Total Direct Effect:
\begin{equation}
    TDE = E\left[Y_{1, Z_1}-Y_{0, Z_1}\right]
\end{equation}
Total Indirect Effect:
\begin{equation}
    TIE = E\left[Y_{1, Z_1}-Y_{1, Z_0}\right]
\end{equation}

\begin{figure*}
    \centering
    \includegraphics[width=16cm]{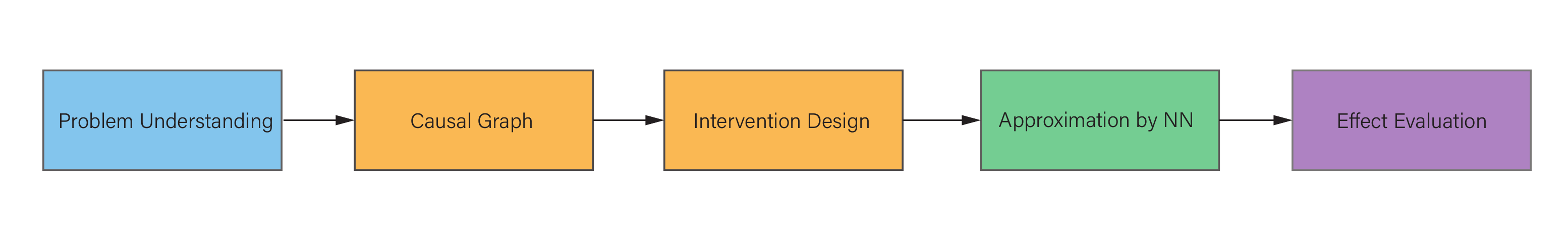}
    \caption{The general workflow for causal representation learning in vision.}
    \label{workflow}
\end{figure*}

\section{Causal Representation Learning in Vision}
In this section, we introduce recent advances of CRL in theoretical works and practical works.

\subsection{Theoretical Work}
Traditional machine learning methods attempt to minimize the empirical risk (ERM). However, these methods can not be generalized to the unseen domain. Moreover, ERM is often over-parameterized, resulting in the discovery of spurious correlations. Based on the concept of causal inference, invariant risk minimization\cite{d1-irm1} is introduced to solve the problems above. Consider a data representation $\Phi: \mathcal{X} \rightarrow \mathcal{H}$. we desire an invariant predictor across the environment $w \circ \Phi$. If there exists a predictor $w: \mathcal{H} \rightarrow \mathcal{Y}$ achieve optimal performance in every environment $\mathcal{E}_{\mathrm{tr}}$ such that $w \in \underset{\bar{w}: \mathcal{H} \rightarrow \mathcal{Y}}{\arg \min } R^e(\bar{w} \circ \Phi)$. Therefore, the constrained optimization problem is defined as:
\begin{equation}
    \begin{array}{ll}
\min _{\substack{\Phi: \mathcal{X} \rightarrow \mathcal{H} \\
w: \mathcal{H} \rightarrow \mathcal{Y}}} & \sum_{e \in \mathcal{E}_{\mathrm{tr}}} R^e(w \circ \Phi) \\
\text { subject to } & w \in \underset{\bar{w}: \mathcal{H} \rightarrow \mathcal{Y}}{\arg \min } R^e(\bar{w} \circ \Phi), \text { for all } e \in \mathcal{E}_{\mathrm{tr}}
\end{array}
\end{equation}
To solve it via gradient descent method, an alternative penalty term is proposed,  
\begin{equation}
    \min _{\Phi: \mathcal{X} \rightarrow \mathcal{Y}} \sum_{e \in \mathcal{E}_{\mathrm{tr}}} R^e(\Phi)+\lambda \cdot\left\|\nabla_{w \mid w=1.0} R^e(w \cdot \Phi)\right\|^2
\end{equation}
where $\mathrm{D}$ measures the risk when changing the environment,
$\lambda \in[0, \infty)$ is a hyper-parameter balancing the ERM and the IRM. This paper only discussed the linear condition of $w$. In the colored MNIST synthetic task, the IRM method achieved 66.9$\%$ accuracy, whereas the ERM method could only achieve 17.1$\%$ accuracy, even lower than random guessing.

Despite the excellent performance on linear conditions, \cite{d1-irm2} proved that IRM can not find the optimal invariant predictor on most occasions and even suffer from a catastrophic failure in a particular condition. This paper first introduced and analyzed the non-linear scene. \cite{d1-irm3} deeply discussed the limitation of the IRM and proved that the IRM prefers an invariant predictor with worse o.o.d generalization. Moreover, the invariant loses its effect when using IRM on empirical samples rather than the population distributions. To address the problem that the IRM fails in the non-linear condition, \cite{d1-irm4} utilized the Bayesian inference to diminish the over-fitting problem and tricks like ELBO and reparameterization to accelerate convergence speed. \cite{d1-irm6} proposed sparse IRM to prevent the spurious correlation leaking to sub-models. On the dataset side, \cite{d1-irm5} proposed a Heterogeneous Risk Minimization (HRM) structure to address the mixture of multi-source data without a source label. Based on this work, \cite{d1-irm8} extended to kernel space, enhancing the ability to deal with more complex data and invariant relationships. \cite{d1-irm7} thought that the IRM method could not deal with the relationship between input and output in different domains. Therefore, \cite{d1-irm7} tried to modify the parameters to make the model robust when domain shifting based on mate-learning structure.

IRM method also mutually benefited from another theory. \cite{li2022invariant} combined information bottleneck with IRM for domain adaptation. The loss is designed in a mutual information expression, whose structure is similar to the IRM.

\begin{table*}[]
\centering
\begin{tabular}{@{}cccc@{}}
\toprule
Task                                                                  & \multicolumn{3}{c}{Performance Improvement over 2nd Model}                             \\ \midrule
\begin{tabular}[c]{@{}c@{}}Scene Graph \\ Generation\cite{tang2020unbiased} \end{tabular}     & Predicate Classification         & Scene Graph Classification  & Scene Graph Detection \\
                                                                      & 51.1\%                           & 56.4\%                      & 31.3\%                \\
\multicolumn{1}{l}{}                                                  & Zero-Shot Relationship Retrieval & Sentence-to-Graph Retrieval & \multicolumn{1}{l}{}  \\
\multicolumn{1}{l}{}                                                  & 25.0\%                           & 33.7\%                      & \multicolumn{1}{l}{}  \\ \midrule
\begin{tabular}[c]{@{}c@{}}Image\\ Captioning\cite{f8}\end{tabular}            & Karpathy Split\cite{karpathy2015deep} 5 Captions        & Karpathy Split Whole Set    & MS-COCO\cite{lin2014microsoft}               \\
                                                                      & 0.250\%                          & 1.55\%                      & 1.02\%                \\ \midrule
\begin{tabular}[c]{@{}c@{}}Attention\\ Mechanism\cite{f10}\end{tabular}       & CNN-Based NICO\cite{he2019nico}                   & CNN-Based ImageNet-9\cite{deng2009imagenet}        & CNN-Based ImageNet-A  \\
                                                                      & 8.72\%                           & 1.15\%                      & 8.33\%                \\
\multicolumn{1}{l}{}                                                  & ViT-Based NICO                   & ViT-Based ImageNet-9        & ViT-Based ImageNet-A  \\
\multicolumn{1}{l}{}                                                  & 8.08\%                           & 2.78\%                      & 12.7\%                \\ \midrule
\begin{tabular}[c]{@{}c@{}}Few-shot \\ Learning\cite{f14}\end{tabular}          & miniImageNet                     & tieredImageNet              &                       \\
                                                                      & 2.40\%                           & 0.94\%                      &                       \\ \midrule
\begin{tabular}[c]{@{}c@{}}Long-tailed \\ Classification\cite{f13}\end{tabular} & LVIS V1.0\cite{gupta2019lvis} val set                & ImageNet-LT                 & LVIS V0.5 val set     \\
                                                                      & 29.2\%                           & 17.3\%                      & 19.1\%                \\
\multicolumn{1}{l}{}                                                  & LVIS V0.5 eval test server       & \multicolumn{1}{l}{}        & \multicolumn{1}{l}{}  \\
\multicolumn{1}{l}{}                                                  & 18.8\%                           & \multicolumn{1}{l}{}        & \multicolumn{1}{l}{}  \\ \midrule
\begin{tabular}[c]{@{}c@{}}Incremental \\ Learning\cite{f12}\end{tabular}       & CIFAR-100\cite{cifar100}                       & ImageNet-Sub                & ImageNet-Full         \\
                                                                      & 6.17\%                           & 4.76\%                      & 3.49\%                \\ \midrule
\begin{tabular}[c]{@{}c@{}}Image \\ Recognition\cite{wang2020visual}\end{tabular}          & MS-COCO                          & Open Images\cite{kuznetsova2020open}                 & VQA2.0 test\cite{antol2015vqa}          \\
                                                                      & 1.41\%                           & 1.09\%                      & 0.560\%               \\ \midrule
\begin{tabular}[c]{@{}c@{}}Visual\\ Grounding\cite{f5}\end{tabular}            & RefCOCO\cite{kazemzadeh2014referitgame}                         & RefCOCO+                    & RefCOCOg              \\
                                                                      & 2.09\%                           & 2.36\%                      & 1.77\%                \\
\multicolumn{1}{l}{}                                                  & ReferIt Game                     & Flickr30K Entities\cite{plummer2015flickr30k}          & \multicolumn{1}{l}{}  \\
\multicolumn{1}{l}{}                                                  & 1.14\%                           & 0.36\%                      & \multicolumn{1}{l}{}  \\ \midrule
\begin{tabular}[c]{@{}c@{}}Weakly-Supervised \\ Learning\cite{f15}\end{tabular} & PASCAL VOC 2012\cite{pascal-voc-2012}                  & MSCOCO                      &                       \\
                                                                      & 1.84\%                           & 2.45\%                      &                       \\ \bottomrule
\end{tabular}
\caption{The performance improvement table for causal representation learning in practical works. Most of the works are plug-and-play and robust to other downstream tasks with only a few costs in total parameters.}
\end{table*}

\subsection{Practical Work}
Feature understanding and transfer learning are two specific research applications in CRL. Confounders are common in vision datasets, which mislead the machine model into catching the bad relationship. Research in feature understanding in CRL attempts to build the SCM and intervene in the node to discover the causal relationship. For transfer learning, statistical methods suffer from the o.o.d data. Discovering the causal or avoiding the adaptation risk reduce the complexity of training a new transfer learning algorithm and improve performance. This section will present the recent CRL advances in feature understanding and transfer learning.

\subsubsection{Feature understanding}
In object detection tasks, the highly correlated objects tend to occur in the same image (e.g. the chair and human, because people could sit on the chair instead of commensalism). VC R-CNN\cite{wang2020visual} thought that observational bias made the model ignore the common causal relationship. They introduced an intervention (confounder dictionary) to measure the true causal effect.  

In scene graph generation, 
\cite{tang2020unbiased} compared the counterfactual scene and factual scene, using a Total Direct Effect (TDE) analysis framework to remove the bias in training. \cite{f9} proposed Align-RCNN to discover the feature relationship and concatenate those features dynamically.

In visual grounding (visual language tasks), the location of the target bounding box highly depends on the query instead of causal reasoning. \cite{f5} proposed a plug-and-play framework Referring Expression Deconfounder (RED), to make the backdoor adjustment to find the causal relationship between images and sentences.

In image captioning (visual language tasks), \cite{f8} thought that the pre-training model contains the confounder. The authors introduced the backdoor adjustment to deconfound the bias. Few-shot learning\cite{wang2020generalizing} also requires the pre-training model. Similarly, \cite{f14} proposed three solutions, including feature-wise adjustment, class-wise adjustment, and class-wise adjustment. These operations do not need to modify the backbone and could be applied easily in zero-shot learning, meta-learning\cite{hospedales2021meta} etc.

In the attention mechanism, most models worked well due to the i.i.d of the data. However, the o.o.d data degrades the performance when using attention. \cite{f10} proposed Casual Attention Module (Caam) on original CBAM-based CNN\cite{woo2018cbam} and ViT\cite{dosovitskiy2020image}. This method utilized the IRM and adversarial training\cite{goodfellow2020generative} with a partitioned dataset to discover confounding factor characteristics.

In feature (disentanglement) representation learning, the classical work simCLR\cite{chen2020simple} proposed a Self-Supervised Learning (SSL), using a contrastive objective method to recognize similar images. However, the generalization performance and the interpretability are poor. \cite{f7} introduced Iterative Partition-based Invariant Risk Minimization (IP-IRM), combining feature disentanglement (data partition) and IRM. This method could discover the critical causal representation for classification tasks. \cite{f11} proposed a counterfactual generation framework for zero-shot learning tasks\cite{larochelle2008zero} based on counterfactual faithfulness theorem. It designed a two-element classifier, disentangling the feature in class and sample.

In incremental learning\cite{masana2020class}, \cite{f12} used causal-effect theory to explain the forget and anti-forget. It designed models for distilling the causal effect of data, colliding effect, and removing SGD momentum\cite{sutskever2013importance} (optimizer) effect of guaranteeing the effectiveness of introducing new data to learn. The interference of the SGD momentum also occurs in long-tailed classification because the update direction of SGD contains the information on data distribution. \cite{f13} designed a multi-head normalized classifier in training and made counterfactual TDE inference in testing to remove the excessive tendency for head class.

In weakly-supervised learning\cite{zhou2018brief} for semantic segmentation, the problems often occur in pseudo-masks, for example, object ambiguity, incomplete background, and incomplete foreground. \cite{f15} introduced a causal intervention------blocking the connection between context prior and images to remove the fake association between label and images. \cite{f15} proposed a context adjustment for the unknown context prior, that is, using a class-specific average mask to approximately constructing a confounder set. \cite{chen2022c} proposed a category causality chain and an anatomy-causality chain to solve the ambiguous boundary and co-occurrence problems in medical image segmentation.

\subsubsection{Transfer learning}
One of the popular research directions in transfer learning is domain adaptation. \cite{subbaswamy2020spec} defined invariance specification: a 2-tuple $\langle\mathcal{P}, \mathbf{M}\rangle$, where $\mathcal{P}$ donates graphical representation, $\mathbf{M}$ donates a group of variable (control the data from different environments). If an invariance spec is found, we could get a stable representation such that the model can be applied to the target domain. Similarly, \cite{d2-domain1} considered domain adaptation as an inference problem, constructing a DAG and solving it by Bayesian inference.

In image-to-image translation, \cite{d2-domain2} concluded a DAG and developed an important reweighting-based learning method. This method can automatically select the images and perform translation simultaneously.

In general, transfer learning, \cite{kunzel2018transfer} introduced the new problem of transfer learning for estimating heterogeneous treatment effects and developed several methods (e.g. Y-Learner). \cite{rojas2018invariant} proposed invariant models for transfer learning. \cite{zhang2017transfer} utilized a causal approach to Multi-Armed Bandits in reinforcement learning.\cite{magliacane2018domain} exploited causal inference to predict invariant conditional distribution in domain adaptation without prior knowledge of the causal graph and the type of interventions. 

\section{Future Work on Medical Image Analysis}
Medical image analysis is a high-risk task, significantly requiring an explainable framework so that doctors and patients can rely on the diagnosis. The current works still lack interpretability and are treated as a  black box. Causal representation learning is a promising learning paradigm for medical image analysis. In this section, we mention some prospective research directions.
\subsection{Attention mechanism}
Attention mechanism is widely applied in medical image analysis, and has shown promising results in lots of datasets and tasks\cite{att1}\cite{att2}\cite{att3}\cite{att4}\cite{att5}\cite{att6}, especially for pure attention model (e.g. ViT\cite{dosovitskiy2020image}, Swin-Transformer\cite{liu2021swin}, Swin-UNet\cite{cao2021swin}). The attention mechanism will bring interpretability to the model. For example, in organ segmentation (e.g. cardiac, brain), attention could highlight the features of a region and suppress other noisy parts. However, the attention mechanism suffers from the data distribution shift (CNN-based attention) and the small scale of datasets (Transformer based). \cite{gonccalves2022survey} shown that the results are noisy on some datasets, even for the cases of attention mechanisms. The causal attention module (Caam)\cite{tang2020unbiased} is a promising method to solve this problem without changing the original framework.
\subsection{Deconfounded optimizer}
Most of the works have various settings of the optimizer, and the performance will be damaged if we change to another optimizer. As \cite{f12}\cite{f13} mentioned above, the optimizer (e.g. SGD momentum) can be a confounder in incremental learning and long-tailed classification. The design of a multi-head normalized classifier and counterfactual TDE inference could be a solution in the medical field.
\subsection{Domain adaptation/generalization}
The data distribution shift problem would decrease the performance of the original model due to the various sources from different hospitals. Recently, domain adaptation\cite{da1}\cite{da2}\cite{da3}\cite{da4}\cite{da5}\cite{da6} and domain generalization\cite{dg1}\cite{dg2}\cite{dg3}\cite{dg4}\cite{dg5} techniques increased a great deal of accuracy (from $\approx 10\%$ to $\approx 70\%$ in these tasks. However, the model must be trained again when adding a new source for adaptation, and the tricks for domain generalization heavily depend on data pre-processing. Additionally, the performance ($\approx 70\%$) still can not be trusted and applied in clinical application. The series works of IRM \cite{d1-irm1} provide a new general solution for domain adaptation/generalization problems.
\subsection{Feature disentanglement}
Feature disentanglement attempts to desperate independent feature to make the model explainable. Specifically, classical feature disentanglement methods utilized the Variational Autoencoder\cite{vae} or GAN\cite{goodfellow2020generative} with a restriction in different channels (e.g. minimize mutual information) to disentangle the high-level semantic representation. In domain adaptation, \cite{dis1} aimed to find the domain-specific feature and the domain invariant feature to make the model robust. In multi-task learning, \cite{dis2} proposed a dual-stream network to share the common feature in latent space. \cite{dis3} utilized a VAE to learn a multi-channel spatial representation of the anatomy. However, the restriction for disentanglement is still loose and lacks interpretability in the current work. We could refer to the concept of IP-IRM\cite{f7} to discover the causal representation of medical images.
\subsection{Class imbalance learning}
The solution to the class imbalance problem traditionally relies on the data pre-processing (e.g. oversampling\cite{kotsiantis2006handling}, re-weighting\cite{zhang2021learning} etc.). But, we can not know the data distribution before training. Additionally, the trick, like re-weighting, will lead the head categories under-fitted. We could refer to \cite{f13} to invent a multi-head normalized classifier in training and make counterfactual TDE inference in testing to solve the long-tailed problem.
\section{Conclusion and Prospect}
This paper reviewed the development of causal representation learning from concept to application. Firstly, we introduce the basic knowledge of causal inference. Secondly, we analyze the theoretical works on IRM and practical works on feature understanding and transfer learning. The existing method shows promising results on benchmark datasets (the performance upgraded over 3-5$\%$ in many areas). Most of the works will not increase the complexity or parameter of the model (with a simple intervention or adjustment), but they are very effective and robust in different tasks. Finally, we also mention some future research directions in general CRL as follows:

\begin{itemize}
    \item A well-defined causal graph is essential. However, some scenes, like anomaly detection, are very complicated. The performance of causal inference is sensitive to the causal graph.
    \item The suitable approximation of operation (intervention, backdoor/frontdoor adjustment) on a node in the causal graph is hard to design. 
    \item Although causal inference is a promising way toward an explainable AI, a complete theory is required with mathematics definition.
    \item We should explore the effect of causal representation learning in downstream applications or tasks (e.g. medical image analysis, low-level vision, etc.)
\end{itemize}

\bibliographystyle{IEEEtran}
\bibliography{ijcai19.bib}

\end{document}